\documentclass[10pt,twocolumn,letterpaper]{article}

\usepackage{cvpr}              

\usepackage[accsupp]{axessibility}
\usepackage{graphicx}
\usepackage{amsmath}
\usepackage{amssymb}
\usepackage{booktabs}
\usepackage[table,xcdraw]{xcolor}
\usepackage{tabularx}
\usepackage[accsupp]{axessibility}  

\usepackage[pagebackref,breaklinks,colorlinks]{hyperref}

\usepackage[capitalize]{cleveref}
\crefname{section}{Sec.}{Secs.}
\Crefname{section}{Section}{Sections}
\Crefname{table}{Table}{Tables}
\crefname{table}{Tab.}{Tabs.}

\newcommand{\srgb}{{\scriptscriptstyle\mathrm{sRGB}}}
\newcommand{\prophoto}{{\scriptscriptstyle\mathrm{PP}}}
\newcommand{\clippedProPhoto}{{\scriptscriptstyle\mathrm{ClippedPP}}}
\newcommand{\image}{{\mathrm{\textbf{I}}}}
\newcommand{\coord}{{\mathrm{\textbf{x}}}}
\newcommand{\imageSRGB}{\image_{\srgb}}
\newcommand{\imageProPhoto}{\image_{\prophoto}}
\newcommand{\imagePredictedProPhoto}{\hat{\image}_{\prophoto}}
\newcommand{\imageClippedProPhoto}{\image_{\clippedProPhoto}}
\newcommand{\R}{\mathbb{R}}
\newcommand{\M}{\mathbf{M}}
\newcommand{\C}{\mathbf{C}}

\def\cvprPaperID{2835} 
\def\confName{CVPR}
\def\confYear{2023}

\begin{document}

\title{GamutMLP: A Lightweight MLP for Color Loss Recovery}

\author{Hoang M. Le$^1$ \hspace{1cm} Brian Price$^2$ \hspace{1cm} Scott Cohen$^2$ \hspace{1cm} Michael S. Brown$^1$\\
$^1$York University \hspace{1.5cm} $^2$Adobe Research\\
\small{
\texttt{\{hminle,mbrown\}@yorku.ca, \{bprice,scohen\}@adobe.com}}
}
\maketitle

\begin{abstract}
Cameras and image-editing software often process images in the wide-gamut ProPhoto color space, encompassing 90\% of all visible colors. However, when images are encoded for sharing, this color-rich representation is transformed and clipped to fit within the small-gamut standard RGB (sRGB) color space, representing only 30\% of visible colors. Recovering the lost color information is challenging due to the clipping procedure. Inspired by neural implicit representations for 2D images, we propose a method that optimizes a lightweight multi-layer-perceptron (MLP) model during the gamut reduction step to predict the clipped values.  {\it GamutMLP} takes approximately 2 seconds to optimize and requires only 23 KB of storage.  The small memory footprint allows our GamutMLP model to be saved as metadata in the sRGB image---the model can be extracted when needed to restore wide-gamut color values. We demonstrate the effectiveness of our approach for color recovery and compare it with alternative strategies, including pre-trained DNN-based gamut expansion networks and other implicit neural representation methods.  As part of this effort, we introduce a new color gamut dataset of 2200 wide-gamut/small-gamut images for training and testing.\vspace{-4mm}
\end{abstract}


\section{Introduction}

\begin{figure}[t!]
  \includegraphics[width=1\linewidth]{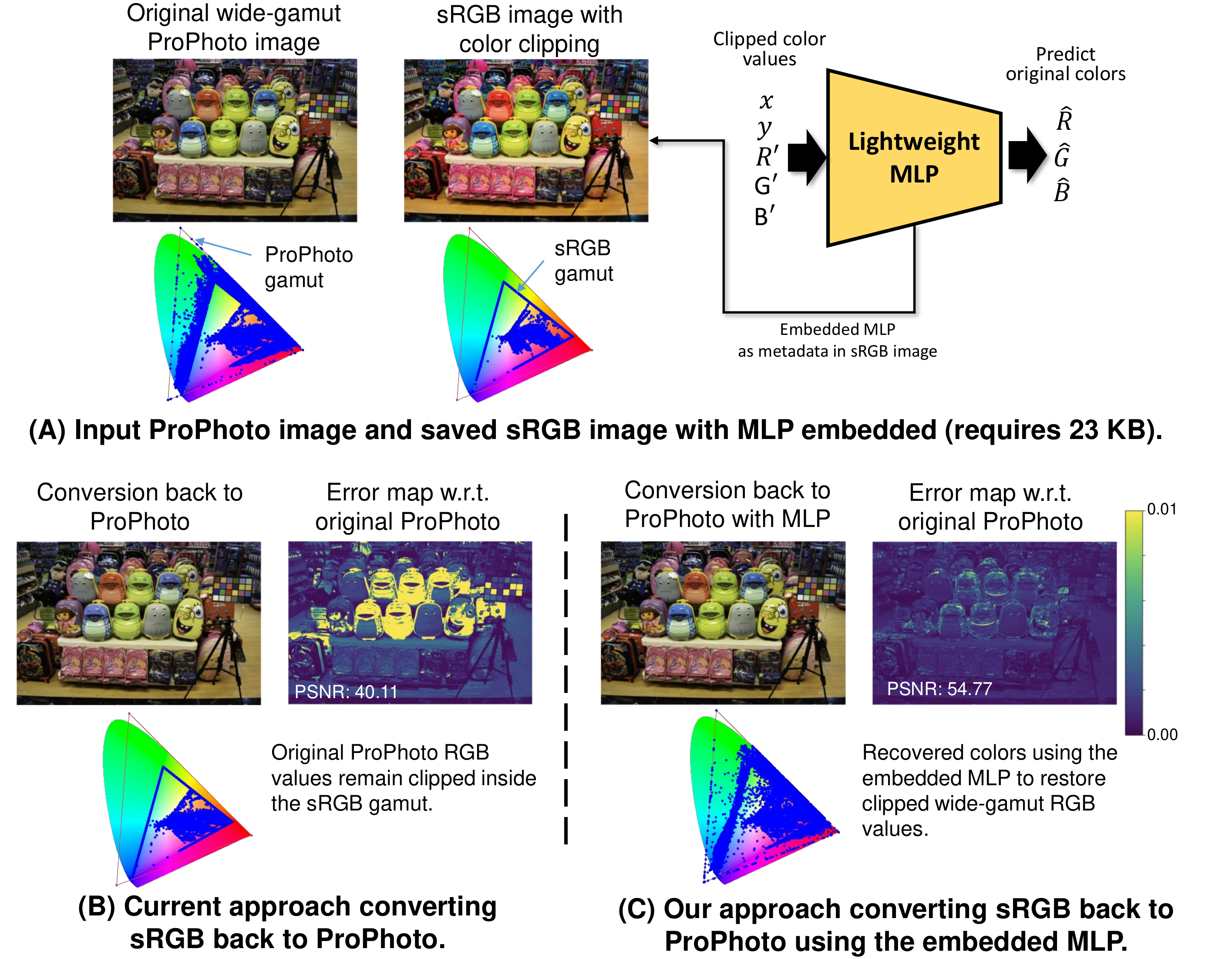}
  \centering
  \vspace{-2mm}
  \caption{(A) shows a wide-gamut (ProPhoto) image that has been converted and saved as a small-gamut (sRGB) image; color clipping is required to fit the smaller sRGB gamut (as shown in the chromaticity diagrams). (B) Standard color conversion back to the wide-gamut color space is not able to recover the clipped colors. (C) Conversion back to the wide-gamut RGB using our lightweight GamutMLP (23 KB) can recover the clipped color values back to their original values.}
  \vspace{-4mm}
  \label{fig:teaser_fig}
\end{figure}

\begin{figure*}[t]
  \includegraphics[width=0.99\textwidth]{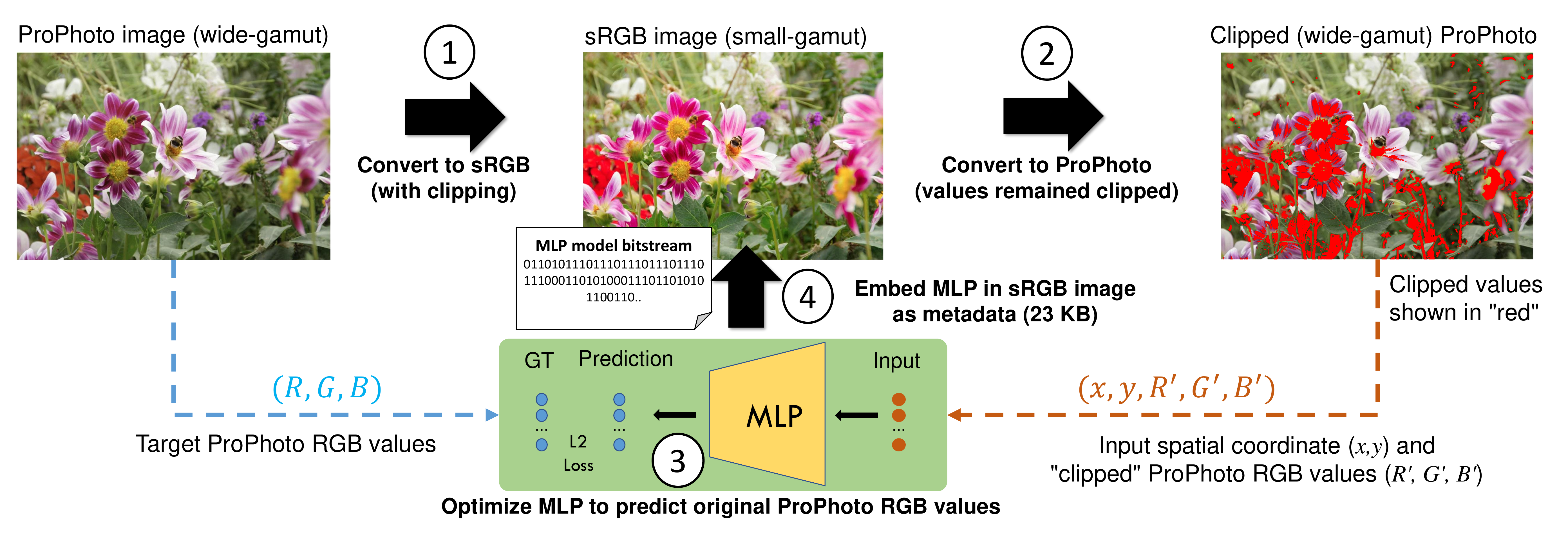}
  \centering
  \vspace{-2.5mm}
  \caption{An overview of the {\it gamut reduction} stage in our framework. This phase shows the gamut reduction step, where the wide-gamut ProPhoto is converted to the small-gamut sRGB. While saving the sRGB image, an MLP is optimized based on the original and clipped ProPhoto color values. The MLP is embedded in the sRGB image as metadata.}
  \vspace{-4mm}
  \label{fig:gamut_reduction_step}
\end{figure*}

\label{sec:intro}
The RGB values of our color images do not represent the entire range of visible colors.  The span of visible colors that can be reproduced by a particular color space's RGB primaries is called a gamut. Currently, the vast majority of color images are encoded using the standard RGB (sRGB) color space~\cite{sRGB}. The sRGB gamut is capable of reproducing approximately 30\% of the visible colors and was optimized for the display hardware of the 1990s. Close to 30 years later, this small-gamut color space still dominates how images are saved, even though modern display hardware is capable of much wider gamuts.

Interestingly, most modern DSLR and smartphone cameras internally encode images using the ProPhoto color space~\cite{karaimer2016software}. ProPhoto RGB primaries define a wide gamut capable of representing 90\% of all visible colors~\cite{prophotorgb}. Image-processing software, such as Adobe Photoshop,  also uses this color-rich space to manipulate images, especially when processing camera RAW-DNG files. By processing images in the wide-gamut ProPhoto space, cameras and editing software allow users the option to save an image in other color spaces---such as AdobeRGB, UHD, and Display-P3---that have much wider color gamuts than sRGB. However, these color spaces are still rare, and most images are ultimately saved in sRGB. To convert color values between ProPhoto and sRGB, a gamut reduction step is applied that clips the wide-gamut color values to fit the smaller sRGB color gamut. Once gamut reduction is applied, it is challenging to recover the original wide-gamut values. As a result, when images are converted back to a wide-gamut color space for editing or display, much of the color fidelity is lost, as shown in Figure~\ref{fig:teaser_fig}.
\\
\noindent{\textbf{Contribution}}~We address the problem of recovering the RGB colors in sRGB images back to their original wide-gamut RGB representation. Our work is inspired by coordinate-based implicit neural image representations that use multilayer perceptrons (MLPs) as a differentiable image representation. We propose to optimize a lightweight (23 KB) MLP model that takes the gamut-reduced RGB values and their spatial coordinates as input and predicts the original wide-gamut RGB values. The idea is to optimize the MLP model when the ProPhoto image is saved to sRGB and embed the MLP model parameters in the sRGB image as a comment field. The lightweight MLP model is extracted and used to recover the wide-gamut color values when needed. We describe an optimization process for the MLP that requires $\sim$2 seconds per full-sized image.  We demonstrate the effectiveness of our method against several different approaches, including other neural image representations and pre-trained deep-learning-based models. As part of this work, we have created a dataset of 2200 wide-gamut/small-gamut image pairs for training and testing.

\section{Related work} \label{sec:related-work}
\begin{figure*}[t]
  \includegraphics[width=0.99\textwidth]{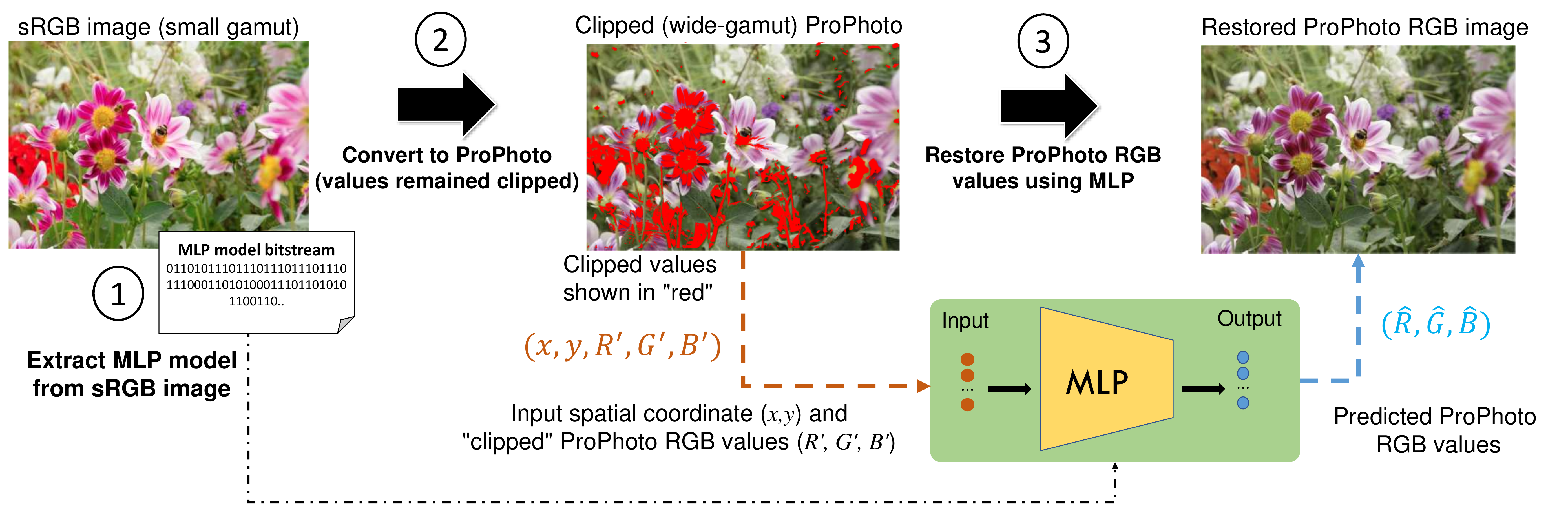}
  \vspace{-2.5mm}
  \centering
  \caption{This figure provides an overview of the {\it gamut expansion} phase in our framework. In particular, the MLP network is extracted and used to recover the clipped ProPhoto values.}
  \vspace{-4mm}
  \label{fig:gamut_expansion_step}
\end{figure*}
The following discusses three areas related to our work: (1) gamut reduction and expansion, (2) RAW image recovery methods, and (3) coordinate-based implicit neural functions.

\noindent{\textbf{Gamut reduction/expansion}}~When converting between color spaces, it is necessary to address the gamut mismatch. There are many strategies for gamut reduction and expansion in the literature (e.g., \cite{JanMorovic_GMA_book,GamutMapping1,mccann2000color,bakke2010evaluation,preiss2014gamut}). The most common approach for both gamut-reduction and gamut-expansion uses {\it absolute colorimetric} intent, where the goal is to minimize color distortion between the two gamuts. For example, in the case of gamut reduction from ProPhoto to sRGB, colorimetric errors are minimized by projecting (and clipping) out-of-gamut (OG) ProPhoto color values to the boundary of the sRGB space, as shown in Figure~\ref{fig:teaser_fig}. When  converting back from sRGB to ProPhoto, the absolute colorimetric strategy minimizes color error by leaving the clipped values untouched. Absolute colorimetric reduction and expansion is a common strategy used by consumer cameras and image-editing software, such as Adobe Lightroom, DarkTable, and RawTherapee. Correcting the loss of color fidelity for color expansion is the goal of this paper.
A less common approach for reduction and expansion is to use soft-clipping~\cite{JanMorovic_GMA_book}, where the out-of-gamut values in ProPhoto are compressed to fit within a specified region in the sRGB gamut. For example, instead of clipping out-of-gamut ProPhoto values, they are compressed to fit within the outer 10\% of the sRGB gamut. On gamut expansion, the compressed region is expanded back to fill the ProPhoto gamut.   While soft-clipping helps restore wide-gamut color values, it incurs colorimetric error during the gamut reduction to sRGB; as a result, most cameras and software do not use this.
Recent work by Le et al. \cite{Hoang_CIC2021} proposed a DNN-based network to learn gamut-expansion based on a large dataset of ProPhoto and sRGB images.  While this can improve gamut expansion, we show that our MLP optimization outperforms such pre-trained DNN networks by a wide margin.

\noindent{\textbf{RAW recovery/derendering}}
Also related to our task are approaches for sRGB {\it derendering}, where the goal is to recover the original RAW sensor image from an sRGB input. Early approaches to this problem carefully modeled the in-camera rendering process to perform derendering~\cite{chakrabarti2014modeling,Chakrabarti2009empirical,kim2012new,gongcic}, while recent methods train DNN-based models for this task~\cite{nam2017modelling,hdrisp}.  Our method is similar to works that save small amounts of specialized metadata in the sRGB image to assist in the recovery problem. Such metadata can be in the form of a parametric model~\cite{nguyen2018raw,rang} or RAW pixel samples~\cite{Punnappurath_2021_WACV,nam_cpvr_2022}. While having access to a reconstructed RAW image would allow it to be re-rendered back to a wide-gamut ProPhoto format, such rendering requires camera-specific photo-finishing parameters that are often not readily available. Instead, we compare our method with that of ~\cite{Hoang_CIC2020}, which proposed to store out-of-gamut ProPhoto samples as metadata in the sRGB. The out-of-gamut samples were used to estimate a polynomial function for gamut expansion. We show that our MLP-based approach provides significantly better results than those obtained by \cite{Hoang_CIC2020} with a smaller memory footprint.

\noindent{\textbf{Neural implicit functions}}
Finally, our approach is also inspired by recent advances in neural implicit functions, such as Fourier features \cite{Tancik:2020:FFN},  SIREN \cite{Sitzmann:2020:SIREN}, and NeRF \cite{NeRF_ECCV2020}.  We show that SIREN works well for this task, but requires a much larger model and slower optimization time. 


\begin{figure*}[t]
  \includegraphics[width=0.8\textwidth]{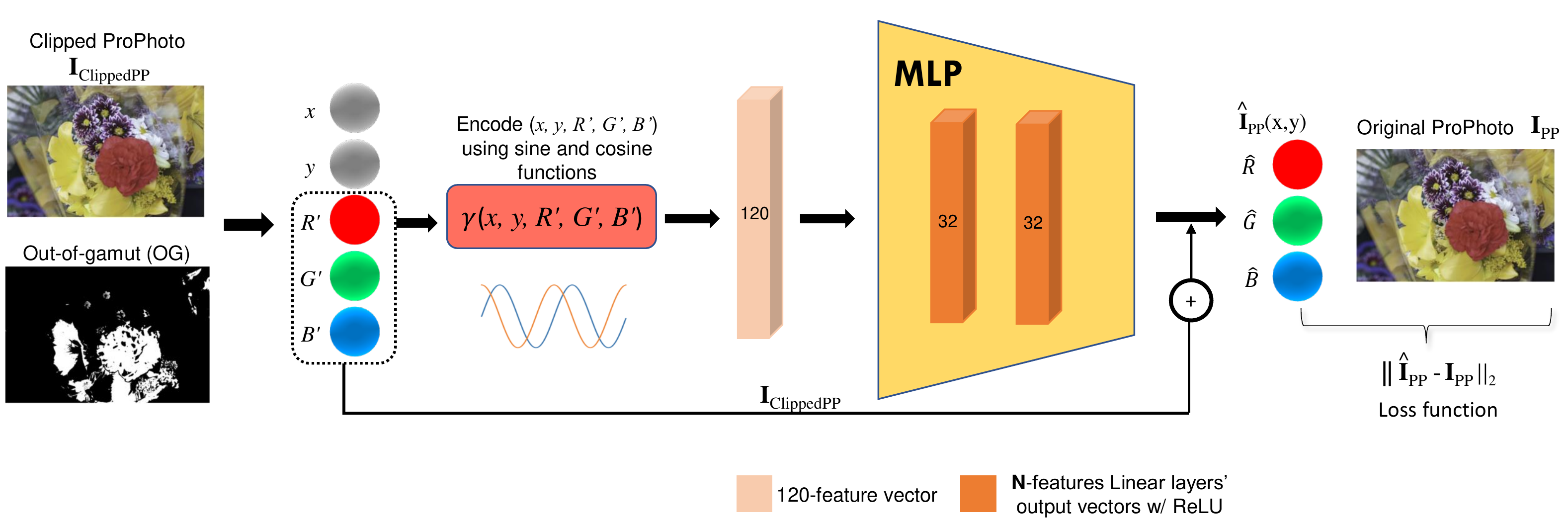}
  \centering
  \vspace{-2.5mm}
  \caption{This figure shows the GamutMLP architecture.  Given the clipped ProPhoto image, we optimize the MLP using samples from in-gamut and out-of-gamut pixels.  The 5D coordinate and color input $(x,y,R',G',B')$ is encoded as a 120D-feature vector before passing it to the MLP. The MLP has three linear layers, with the final layer predicting a residual to add to the $R', G', B'$ input.  The loss is computed against the original ProPhoto image $R, G, B$ values.}
  \vspace{-4mm}
  \label{fig:mlp-overview}
\end{figure*}

\section{GamutMLP for color recovery}

We begin with a high-level overview of our gamut recovery framework in Section~\ref{sec:overview}. Details on the MLP architecture and optimization are provided in Section~\ref{sec:optimize}.

\subsection{Framework overview}\label{sec:overview}

Figure~\ref{fig:gamut_reduction_step} and Figure~\ref{fig:gamut_expansion_step} show illustrations of our framework's two steps: gamut reduction and gamut expansion.
\\
\noindent{\textbf{Gamut reduction}}~~The gamut reduction step is performed when the image is being converted from ProPhoto to sRGB, either on a camera or image editing software.  We assume the input to be a wide-gamut ProPhoto RGB image denoted as $\imageProPhoto \in \R^{3 \times N}$, where $N$ is the number of pixels. Gamut reduction is performed using the absolute colorimetric intent described in the previous section.  The original ProPhoto image is transformed to the unclipped sRGB image using a $3\times3$ matrix such that the in-gamut sRGB values fall within the range $[0,1]$. Out-of-gamut sRGB values are then clipped and processed with a gamma encoding to produce the final sRGB image.  This procedure can be written as:
\begin{equation} \label{eq:1}
    \imageSRGB = g(\textrm{clip}(\M \imageProPhoto, min=0, max=1)),
\end{equation}
where $\M$ is the matrix that maps between ProPhoto and the unclipped sRGB, $\textrm{clip}()$ is the clipping operation, and $g$ is the gamma-encoding for sRGB~\cite{sRGB}.

When converting sRGB colors back to ProPhoto RGB using the inverse transforms, the clipped color values will not be recovered. We refer to this clipped ProPhoto image as $\imageClippedProPhoto \in \R^{3 \times N}$. Pixels with clipped values are illustrated in red in Figure~\ref{fig:gamut_reduction_step}.  We express the mapping from sRGB to clipped ProPhoto as:
\begin{equation} \label{eq:2}
    \imageClippedProPhoto = \M^{-1} g^{-1}(\imageSRGB),
\end{equation}
where $g^{-1}(\cdot)$ is a de-gamma function for the input sRGB image, and $\M^{-1}$ is the inverse transform to convert the sRGB image back to the ProPhoto color space.

Applying Eqs.~\ref{eq:1} and~\ref{eq:2}, we have the original ProPhoto image, $\imageProPhoto$, and its clipped $\imageClippedProPhoto$.  We also know which values were clipped.  Using these two images, we optimize a lightweight MLP (GamutMLP) to predict a residual value that, when added to the  $\imageClippedProPhoto$ recovers $\imageProPhoto$. The GamutMLP model parameters are embedded in the sRGB image when it is saved.  Since the parameters of our MLP require only 23 KB of memory, this can easily be embedded as a comment field in the image.

\noindent{\textbf{Gamut expansion step}}~~Given an sRGB image with an embedded GamutMLP model, we extract the model and perform the standard color space conversion described in Eq.~\ref{eq:2} to compute $\imageClippedProPhoto$. The extracted model predicts the residuals of all pixels and adds them to the $\imageClippedProPhoto$ to recover the color values as shown in Figure~\ref{fig:gamut_expansion_step}.

The following section provides details of the GamutMLP model architecture and its optimization. 

\subsection{GamutMLP model and optimization}\label{sec:optimize}

Figure~\ref{fig:mlp-overview} shows a diagram of the GamutMLP architecture.  When converting from ProPhoto to sRGB, we keep track of the transformed RGB values that lie outside the sRGB gamut.  These  pixel locations are denoted in the out-of-gamut mask in Figure~\ref{fig:mlp-overview}. Out-of-gamut pixels will be clipped to fit within the sRGB gamut.  

The MLP input is a 5D vector of a pixel's spatial coordinates $(x,y)$ and color value $(R', B', G')$ from the clipped wide-gamut ProPhoto image. GamutMLP predicts the residual that needs to be added to $\imageClippedProPhoto$ to recover the wide-gamut original $(R, G, B$).  We can express GamutMLP as follows:
\begin{equation}
    \imagePredictedProPhoto(\coord) = f_{\theta}(\coord, \imageClippedProPhoto(\coord)) + \imageClippedProPhoto(\coord),
\end{equation}
where $f_\theta$ represents the \textit{GamutMLP}, $\theta$ is the model's parameters, and $\imagePredictedProPhoto$ is the final recovered ProPhoto image.  The MLP's input values $(x, y, R', G', B')$ are normalized to the range $[-1, 1]$, and then pass to the encoding function $\gamma$. The use of an encoding function has been shown effective in improving neural implicit representations optimization~\cite{NeRF_ECCV2020,rahaman19a}. Prior neural implicit representations tend to have only 2D coordinates as input, while we apply the encoding function to both spatial coordinates and RGB values. 
We found the following mapping worked well for our task:
\begin{equation}\label{eq:encoding}
\resizebox{0.9\hsize}{!}{
    $\gamma(m) = (\sin(2^{0}\pi m),\cos(2^{0}\pi m), ...,\sin(2^{K-1}\pi m),\cos(2^{K-1}\pi m)),$
    }
\end{equation}
where $m$ is a spatial coordinate or RGB value. In our experiment, we choose K = 12.  The $\gamma$ function projects each of the 5D input values to a 24-dimension encoding, resulting in a final 120D feature vector for each input.  The GamutMLP has three linear layers. The first two are fully connected ReLU layers with 32 output features. The last layer outputs three values and has no activation function.  Our MLP is optimized with an $L_2$ loss function computed between the predicted ProPhoto image and the original ProPhoto image:
\begin{equation}\label{eq:loss}
    {\cal{L}}_{gamut} = \sum_{\coord}||(\imagePredictedProPhoto(\coord) - \imageProPhoto(\coord))||^2_2.
\end{equation}

\noindent{\textbf{Pixel sampling and standard optimization}}~
We describe two optimization strategies: standard and fast. For our standard optimization, model parameters are randomly initialized. At optimization time, we know which pixels are out-of-gamut (OG) and which are in-gamut (IG).  We found that training the MLP on both out-of-gamut pixels and in-gamut (non-clipped) pixels gave the best result.  We optimized our model by sampling 2\% of the IG pixels and 20\% of the OG pixels uniformly over their spatial coordinates.  For all reported results, the model was optimized for 9,000 iterations with a learning rate of $1e-3$ using the Adam optimizer~\cite{adam}.  


\begin{figure*}[t]
\vspace{-0.2cm}
\includegraphics[width=.95\textwidth]{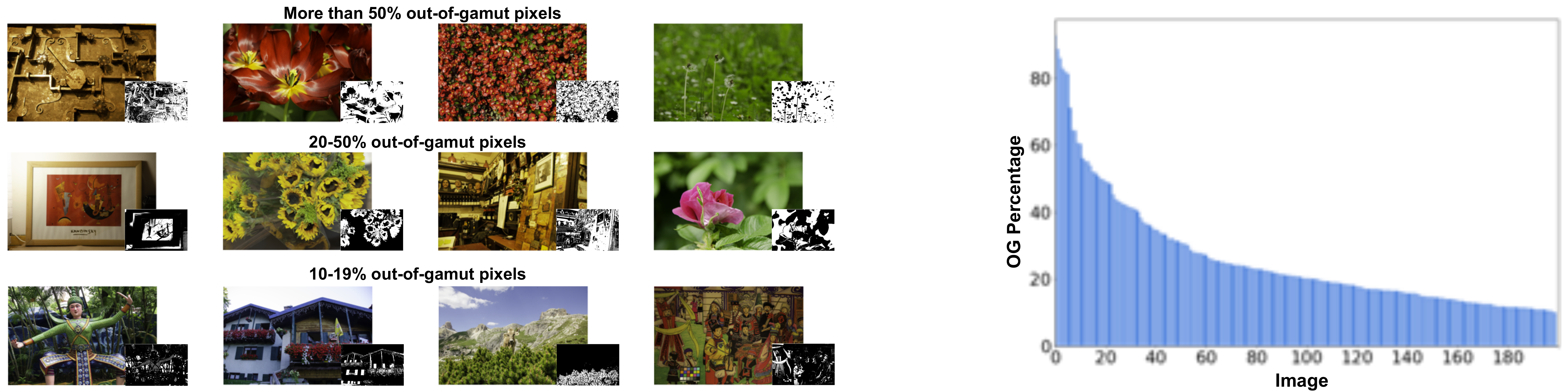}
  \centering
  \vspace{-2.5mm}
  \caption{(Left) Examples from our dataset showing different amounts of out-of-gamut pixels (OG pixels are white in the masks). (Right) A histogram of our dataset in terms of percentage of out-of-gamut pixels.}
  \label{fig:DatasetHistogram}
  \vspace{-4mm}
\end{figure*}

\noindent{\textbf{Faster MLP optimization}}~
To speed up our optimization, we used the recent methods proposed by ~\cite{mueller2021realtime,mueller2022instant} to
improve optimization time.  We also incorporated a meta-learning strategy~\cite{metalearning} to pre-train a generic GamutMLP whose model parameters can be used for initialization. Such initialization has been shown to help coordinate-based MLPs converge faster during optimization~\cite{tancik2021learned}.  

To pre-train a meta-MLP model, we used images from our training dataset described in the following section. For each meta epoch, the meta-MLP fits each image with 10,000 iterations using a larger learning rate $1e-2$ and the SGD optimizer. When a per-image GamutMLP is initialized with this pre-trained MLP, our optimization requires only 1,200 iterations instead of 9,000.  Combining our meta-MLP for initialization with~\cite{mueller2021realtime,mueller2022instant} significantly reduced optimization time.


\section{Dataset and results}

We first describe our dataset generation for evaluating this work.  Our dataset images are used to train our meta-GamutMLP model (for weight initialization) and competing DNN-based methods and to test results.

\subsection{Dataset} \label{sec:dataset}

To prepare wide-gamut ProPhoto images, we followed a procedure used in our prior work~\cite{Hoang_CIC2021} that processed RAW images from the MIT-Adobe FiveK~\cite{fivek}, RAISE~\cite{RAISE}, Cube+~\cite{CUBEPlus}, and NUS~\cite{NUS} public datasets.  There are 16,599 RAW images from these four datasets, representing a wide range of scene content. We use  Adobe Camera RAW (ACR) to mimic a camera ISP to render the RAW images to 16-bit wide-gamut ProPhoto images. ACR can apply different photo-finishing styles when rendering RAW images. We use four picture styles---Adobe Standard, Adobe Landscape, Adobe Color, and Adobe Vivid---to generate ProPhoto images with vivid colors. These are saved in their original full resolution size, ranging from 2000$\times$3000 to 4000$\times$6000.  

From these rendered images, we chose 2000 images for the training set and 200 for the testing set.  Images were selected such that they have at least 10\% out-of-gamut pixels. Figure~\ref{fig:DatasetHistogram} shows a sample of the images from our dataset and a plot of the percentage of out-of-gamut pixels in each image in the dataset.  The minimum number of out-of-gamut pixels for an image in our testing set is over a million pixels. The final breakdown of images selected from the starting datasets is: 11\% Adobe FiveK, 62\% RAISE, 12\% Cube+, and 15\% NUS.

\subsection{Comparisons}
The following describes methods used for comparisons against the GamutMLP approach.

\noindent{\textbf{Conventional methods}}~The two baseline methods are clipping and soft-clipping~\cite{JanMorovic_GMA_book}, described in Section \ref{sec:related-work}. Clipping is currently the de facto method for gamut reduction.  While soft-clipping aids in gamut expansion, it is not commonly used because it distorts the colors in the sRGB.

\noindent{\textbf{Pre-trained deep networks}}~As described in Section \ref{sec:related-work},  GamutNet \cite{Hoang_CIC2021} is a DNN-based method targeting gamut recovery.   For the sake of completeness, we also compare against several image-to-image-translation methods: pix2pix \cite{pix2pix}, pix2pixHD \cite{pix2pixHD}, and ASAPNet \cite{ASAPNet}.  We can consider our problem of clipped-ProPhoto to ProPhoto conversion as a special application for image-to-image translation methods.  For all methods that need to be pre-trained, we train them with  512$\times$512 crops from training images (clipped ProPhoto and ProPhoto pairs). Cropping is performed such that at least 10\% of the cropped image has out-of-gamut pixels.  At inference time, the DNN-based methods are applied to the full-sized testing images.

\noindent{\textbf{Per-image optimization}} We also compare with \cite{Hoang_CIC2020}, another metadata approach that uniformly samples pixels (135KB) of the original ProPhoto image for recovery using polynomial color correction functions. 
We also compare with several variants of the SIREN~\cite{Sitzmann:2020:SIREN} coordinate-based neural implicit function.   In particular, we start with the original SIREN, which uses 2D coordinates for input, and has five fully connected linear layers, each with 256 channels and periodic activation functions; the last linear output layer has only three channels for RGB values. The SIREN model requires 796 KB. We optimize a SIREN-residual variant that predicts the residual between the clipped and ground-truth ProPhoto images.  Finally, we try a variant of SIREN-residual that is limited to a small model size (69 KB) to mimic a smaller model. The SIREN models are optimized based on the loss ${\cal{L}}_{gamut}$ described in Equation~\ref{eq:loss}. For the other hyperparameters, we adopt default settings for image-fitting tasks from SIREN~\cite{Sitzmann:2020:SIREN}. Unfortunately, we could not use the fast implementation of MLP~\cite{mueller2021realtime,mueller2022instant} for the original SIREN and variants since the fast implementation API does not support the sinusoidal activation function used by SIREN. 

For our GamutMLP approach, we show the results of our MLP variants with encoded inputs (i.e., Equation~\ref{eq:encoding}), and with and without optimization plus the meta-GamutMLP initialization.  The pre-trained DNNs methods and MLP-based approaches are trained or optimized using a NVIDIA Quadro RTX 6000.

\begin{table*}[t]
\vspace{-2mm}
\centering
\resizebox{\textwidth}{!}{%
\begin{tabular}{|lcccccc|}
\hline
\multicolumn{1}{|l|}{\textbf{Method}} &
  \multicolumn{1}{c|}{\textbf{Metadata$\downarrow$}} &
  \multicolumn{1}{c|}{\textbf{RMSE$\downarrow$}} &
  \multicolumn{1}{c|}{\textbf{RMSE OG$\downarrow$}} &
  \multicolumn{1}{c|}{\textbf{PSNR$\uparrow$}} &
  \multicolumn{1}{c|}{\textbf{PSNR OG$\uparrow$}} &
  \textbf{Optim. Time$\downarrow$} \\ \hline
\multicolumn{7}{|c|}{\textit{\textbf{Conventional}}} \\ \hline
\multicolumn{1}{|l|}{Clip} &
  \multicolumn{1}{c|}{-} &
  \multicolumn{1}{c|}{0.0069} &
  \multicolumn{1}{c|}{0.0126} &
  \multicolumn{1}{c|}{43.22} &
  \multicolumn{1}{c|}{37.98} &
  - \\ \hline
\multicolumn{1}{|l|}{Soft Clip} &
  \multicolumn{1}{c|}{-} &
  \multicolumn{1}{c|}{0.0039} &
  \multicolumn{1}{c|}{0.0042} &
  \multicolumn{1}{c|}{48.17} &
  \multicolumn{1}{c|}{47.54} &
  - \\ \hline
\multicolumn{7}{|c|}{\textit{\textbf{Pre-trained DNN}}} \\ \hline
\multicolumn{1}{|l|}{Pix2pix\cite{pix2pix}} &
  \multicolumn{1}{c|}{-} &
  \multicolumn{1}{c|}{0.0087} &
  \multicolumn{1}{c|}{0.0167} &
  \multicolumn{1}{c|}{41.24} &
  \multicolumn{1}{c|}{35.55} &
  - \\ \hline
\multicolumn{1}{|l|}{Pix2pixHD\cite{pix2pixHD}} &
  \multicolumn{1}{c|}{-} &
  \multicolumn{1}{c|}{0.0157} &
  \multicolumn{1}{c|}{0.0314} &
  \multicolumn{1}{c|}{36.08} &
  \multicolumn{1}{c|}{30.07} &
  - \\ \hline
\multicolumn{1}{|l|}{ASAPNet\cite{ASAPNet}} &
  \multicolumn{1}{c|}{-} &
  \multicolumn{1}{c|}{0.0518} &
  \multicolumn{1}{c|}{0.0993} &
  \multicolumn{1}{c|}{25.72} &
  \multicolumn{1}{c|}{20.06} &
  - \\ \hline
\multicolumn{1}{|l|}{GamutNet\cite{Hoang_CIC2021}} &
  \multicolumn{1}{c|}{-} &
  \multicolumn{1}{c|}{0.0052} &
  \multicolumn{1}{c|}{0.0088} &
  \multicolumn{1}{c|}{45.75} &
  \multicolumn{1}{c|}{41.08} &
  - \\ \hline
\multicolumn{7}{|c|}{\textit{\textbf{Optimized per image}}} \\ \hline
\multicolumn{1}{|l|}{ProPhoto-Sampled\cite{Hoang_CIC2020}} &
  \multicolumn{1}{c|}{135 KB} &
  \multicolumn{1}{c|}{0.0032} &
  \multicolumn{1}{c|}{0.0051} &
  \multicolumn{1}{c|}{49.78} &
  \multicolumn{1}{c|}{45.90} &
  - \\ \hline
\multicolumn{1}{|l|}{SIREN\cite{Sitzmann:2020:SIREN}} &
  \multicolumn{1}{c|}{796 KB} &
  \multicolumn{1}{c|}{0.0648} &
  \multicolumn{1}{c|}{0.0421} &
  \multicolumn{1}{c|}{23.77} &
  \multicolumn{1}{c|}{27.52} &
  115.67 mins \\ \hline
\multicolumn{1}{|l|}{SIREN-residual} &
  \multicolumn{1}{c|}{796 KB} &
  \multicolumn{1}{c|}{0.0033} &
  \multicolumn{1}{c|}{0.0044} &
  \multicolumn{1}{c|}{49.72} &
  \multicolumn{1}{c|}{47.20} &
  118.98 mins \\ \hline
\multicolumn{1}{|l|}{SIREN   (small)-residual} &
  \multicolumn{1}{c|}{69 KB} &
  \multicolumn{1}{c|}{0.0040} &
  \multicolumn{1}{c|}{0.0052} &
  \multicolumn{1}{c|}{47.98} &
  \multicolumn{1}{c|}{45.66} &
  94.62 mins \\ \hline
\multicolumn{1}{|l|}{MLP + enc. (no optimization)} &
  \multicolumn{1}{c|}{48 KB} &
  \multicolumn{1}{c|}{0.0021} &
  \multicolumn{1}{c|}{0.0031} &
  \multicolumn{1}{c|}{53.57} &
  \multicolumn{1}{c|}{50.17} &
  37.05 sec \\ \hline
\rowcolor[HTML]{FFFC9E}
\multicolumn{1}{|l|}{\cellcolor[HTML]{FFFC9E}MLP (53KB) + enc.} &
  \multicolumn{1}{c|}{\cellcolor[HTML]{FFFC9E}53 KB} &
  \multicolumn{1}{c|}{\cellcolor[HTML]{FFFC9E}0.0021} &
  \multicolumn{1}{c|}{\cellcolor[HTML]{FFFC9E}0.0030} &
  \multicolumn{1}{c|}{\cellcolor[HTML]{FFFC9E}53.73} &
  \multicolumn{1}{c|}{\cellcolor[HTML]{FFFC9E}50.33} &
  16.29 sec \\ \hline
\rowcolor[HTML]{FFFC9E}
\multicolumn{1}{|l|}{\cellcolor[HTML]{FFFC9E}MLP   (23 KB) + enc.} &
  \multicolumn{1}{c|}{\cellcolor[HTML]{FFFC9E}23 KB} &
  \multicolumn{1}{c|}{\cellcolor[HTML]{FFFC9E}0.0021} &
  \multicolumn{1}{c|}{\cellcolor[HTML]{FFFC9E}0.0031} &
  \multicolumn{1}{c|}{\cellcolor[HTML]{FFFC9E}53.65} &
  \multicolumn{1}{c|}{\cellcolor[HTML]{FFFC9E}50.04} &
  16.32 sec \\ \hline
\rowcolor[HTML]{FFFC9E}
\multicolumn{1}{|l|}{\cellcolor[HTML]{FFFC9E}MLP (53KB) +   enc. + meta init.} &
  \multicolumn{1}{c|}{\cellcolor[HTML]{FFFC9E}53 KB} &
  \multicolumn{1}{c|}{\cellcolor[HTML]{FFFC9E}0.0021} &
  \multicolumn{1}{c|}{\cellcolor[HTML]{FFFC9E}0.0032} &
  \multicolumn{1}{c|}{\cellcolor[HTML]{FFFC9E}53.40} &
  \multicolumn{1}{c|}{\cellcolor[HTML]{FFFC9E}50.00} &
  1.94 sec \\ \hline
\rowcolor[HTML]{FFFC9E}
\multicolumn{1}{|l|}{\cellcolor[HTML]{FFFC9E}MLP (23 KB) +   enc. + meta init.} &
  \multicolumn{1}{c|}{\cellcolor[HTML]{FFFC9E}23 KB} &
  \multicolumn{1}{c|}{\cellcolor[HTML]{FFFC9E}0.0021} &
  \multicolumn{1}{c|}{\cellcolor[HTML]{FFFC9E}0.0032} &
  \multicolumn{1}{c|}{\cellcolor[HTML]{FFFC9E}53.36} &
  \multicolumn{1}{c|}{\cellcolor[HTML]{FFFC9E}49.93} &
  1.90 sec \\ \hline
\end{tabular}%
}
\vspace{-2mm}
\caption{This table shows results on various methods used for wide-gamut color recovery. The reported numbers are the average results computed against the 200 16-bit ProPhoto ground-truth full-size images.  RMSE and PSNR are provided for the whole image and out-of-gamut (OG) pixels. For the per-image methods, we provide associated metadata (size in KB) and optimization time.  For pre-trained DNNs, we compare with Pix2pix~\cite{pix2pix}, Pix2PixHD~\cite{pix2pixHD}, ASAPNet~\cite{ASAPNet}, and GamutNet~\cite{Hoang_CIC2021}.  For the per-image methods, we compare with ProPhoto-Sampled~\cite{Hoang_CIC2020}, SIREN~\cite{Sitzmann:2020:SIREN} variants, and our GamutMLP variants: MLP (no optimization), sizes (53 KB) and (23 KB), and with ``meta'' initialization.}
\vspace{-4mm}
\label{tab:fullsize_table}
\end{table*}

\begin{figure*}[ht!]
  \vspace{-0.7cm}
  \includegraphics[width=0.86\textwidth]{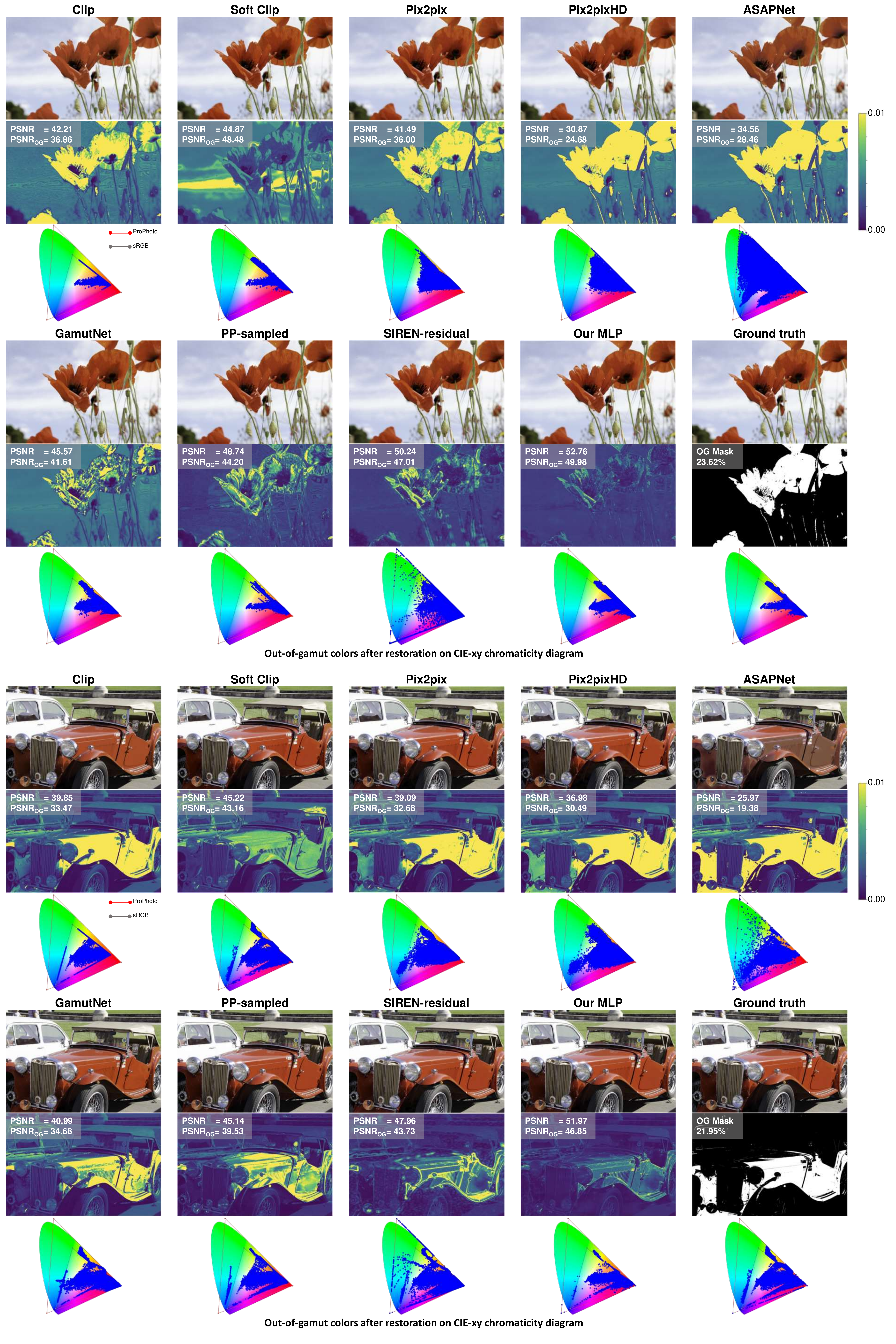}
  \centering
  \vspace{-0.4cm}
  \caption{Qualitative comparisons between the predicted ProPhoto full-size output of Clip, Soft Clip, Pix2pix~\cite{pix2pix}, Pix2PixHD~\cite{pix2pixHD}, ASAPNet~\cite{ASAPNet}, GamutNet~\cite{Hoang_CIC2021}, PP-sampled~\cite{Hoang_CIC2020}, SIREN~\cite{Sitzmann:2020:SIREN}-residual, and our optimized GamutMLP. Error maps of per-pixel RMSE and plots of out-of-gamut (OG) colors on CIE-xy chromaticity diagram with the gamuts of sRGB and ProPhoto are shown.}
  \label{fig:visual-results1}
  \vspace{-0.5cm}
\end{figure*}

\subsection{Quantitative results}

Table~\ref{tab:fullsize_table} summarizes the performance of the tested methods. Results are reported as the average root mean square error (RMSE) and peak signal-to-noise ratio (PSNR) between the predicted and ground truth test images. Metrics are provided for the entire image and for only out-of-gamut (OG) pixels. We provide the associated metadata size (KB) and optimization times for the relevant methods.

The table reveals that simple image-to-image translation does not work well for this task. Pix2pix~\cite{pix2pix} gives better results than Pix2pixHD~\cite{pix2pixHD}, since Pix2pix is a general-purpose method for image-to-image translation, while Pix2pixHD was designed to synthesize images from semantic label maps. The DNN-based GamutNet~\cite{Hoang_CIC2021} method explicitly designed for gamut recovery performs better than the clipping baseline.  The best results are obtained from the per-image optimized methods.  In particular, the full-sized SIREN-residual MLP model produces results similar to ours.  However, the model requires 796KB and is slow to optimize even compared with our equivalent standard optimized MLP. Our GamutMLP with feature encoding and pre-trained initialization gives the best results and the fastest optimization performance. Note that the ProPhoto-sampled method~\cite{Hoang_CIC2020} is optimized per image; however, the computationally intensive component is during the gamut expansion phase instead of the reduction phase. In addition, the method's implementation uses Matlab and runs on CPU, so we do not include its running time since the comparison is unfair.  

\subsection{Qualitative results}
\begin{figure*}[t!]
  \includegraphics[width=.85\textwidth]{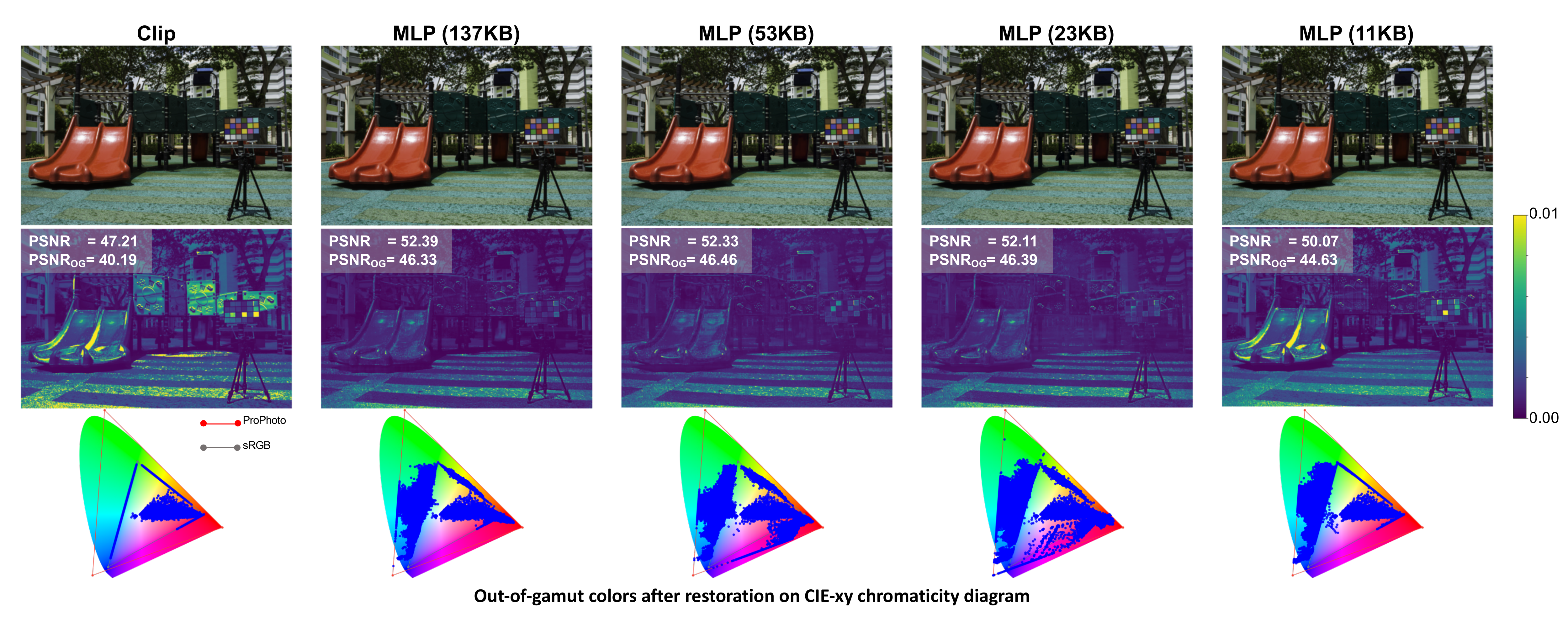}
  \centering
  \vspace{-0.4cm}
  \caption{Qualitative comparisons between the predicted ProPhoto full-size output of Clip and our various versions of MLP (137 KB, 53 KB, 23 KB, and 11 KB).}
  \vspace{-4mm}
  \label{fig:visual-modelsize}
\end{figure*}
We show qualitative visual outputs of selected approaches in Figure~\ref{fig:visual-results1}. See supplemental materials for additional results. For each method, the first row shows the predicted ProPhoto image. The second row shows an error map of the RMSE of each pixel between the predicted and original ProPhoto. The third row shows the out-of-gamut colors after restoration on a CIE-xy chromaticity diagram. Similar to the quantitative results, we see our approach achieves better results in terms of RMSE and PSNR. In addition, the CIE-xy chromaticity diagram shows the recovered colors appear more like the ground truth wide-gamut images. 

While not the goal of our work, on careful observation, it can be noticed that our method and the ProPhoto-sampled method~\cite{Hoang_CIC2020} provide a slight improvement in the in-gamut pixels.  
This is attributed to a slight recovery of some of the 16-bit values that were quantized when the image was saved in 8-bit sRGB in the gamut reduction step.  Recall that this method includes in-gamut pixels in the optimization and is applied by the GamutMLP to all pixels. Similarly, the work \cite{Hoang_CIC2020} uniformly samples ProPhoto pixels, including in-gamut and out-of-gamut pixels.  The most significant improvements, however, are still with the clipped out-of-gamut pixels, which incur the most error in the gamut reduction step.

\subsection{Ablations}

\noindent{\textbf{Input coordinates}}~We trained our proposed MLP with various input types: only coordinates (x,y), only color values (R,G,B), and our 5D vector (x,y,R,G,B) in Table~\ref{tab:mlp_input_types}. The use of the coordinate and color values provides excellent results with a small network. 

\begin{table}[h]
\vspace{-2mm}
\resizebox{\columnwidth}{!}{%
\begin{tabular}{|l|c|c|c|c|}
\hline
\textbf{Method}                 & \multicolumn{1}{l|}{\textbf{RMSE$\downarrow$}} & \multicolumn{1}{l|}{\textbf{RMSE OG$\downarrow$}} & \multicolumn{1}{l|}{\textbf{PSNR$\uparrow$}} & \multicolumn{1}{l|}{\textbf{PSNR OG$\uparrow$}} \\ \hline
MLP (23 KB) [xy]    & 0.0037                             & 0.0048                                & 48.57                              & 46.38                                 \\ \hline
MLP (23 KB) [RGB] & 0.0028                             & 0.0045                                & 51.10                              & 46.98                                 \\ \hline
\rowcolor[HTML]{FFFC9E} 
MLP (23 KB) [xyRGB] & 0.0021                             & 0.0031                                & 53.65                              & 50.04                                 \\ \hline
\end{tabular}
}
\vspace{-2mm}
\caption{This table shows the results of our variant MLPs with various types of inputs: \textit{xy}, \textit{RGB}, \textit{xyRGB}. RMSE and PSNR for the whole image as well as OG pixels are computed and averaged over 200 test images.}
\vspace{-4mm}
\label{tab:mlp_input_types}
\end{table}

\begin{figure}
\includegraphics[width=0.75\columnwidth]{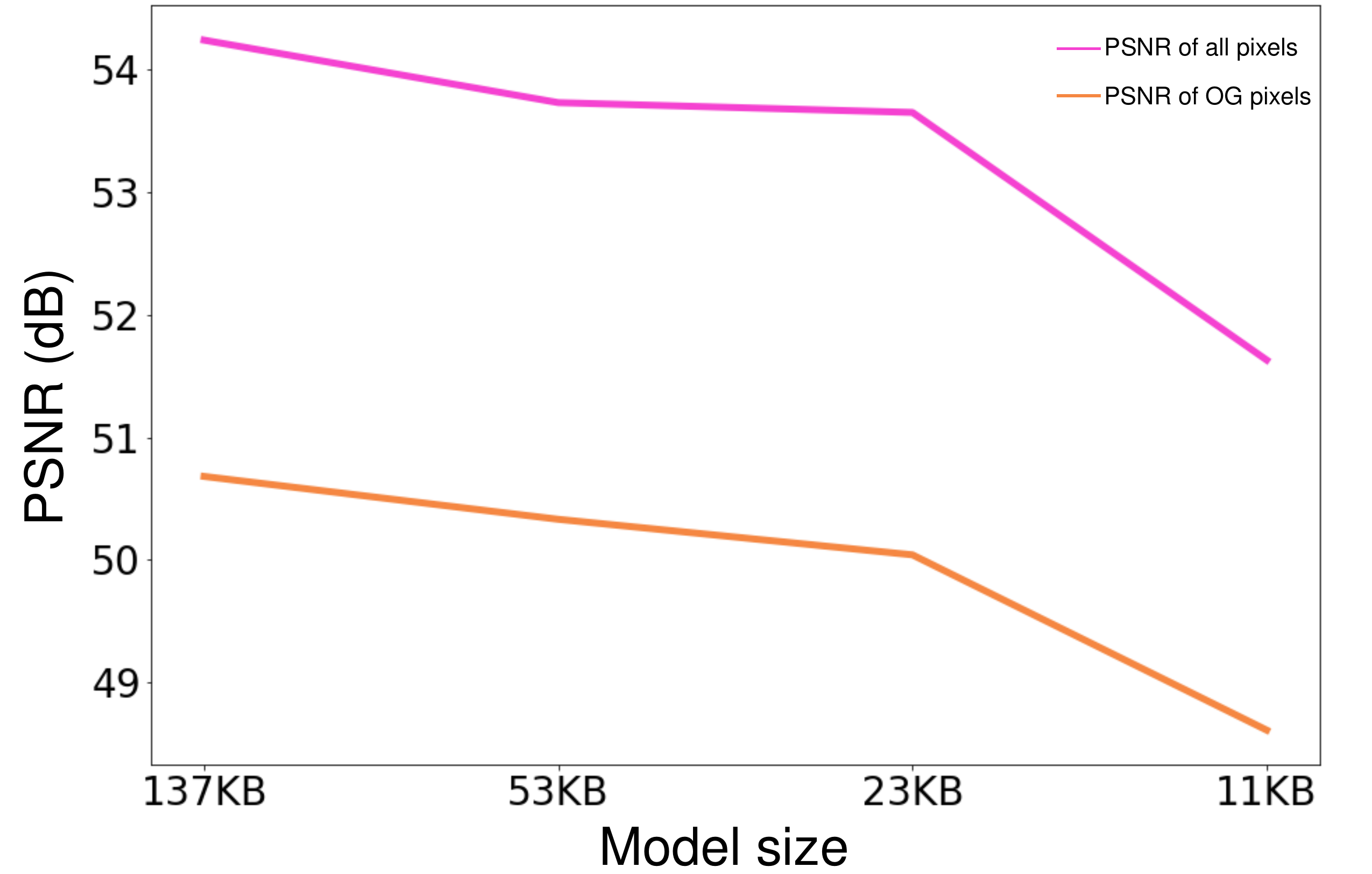}
  \vspace{-3mm}
  \centering
\caption{Ablation examining MLP model sizes.  The 23 KB MLP provided good results for model size.}
\vspace{-4mm}
\label{fig:model_size_PSNR}
\end{figure}

\noindent{\textbf{GamutMLP model sizes}} Our goal was to find a model size that was compact and had a reasonable optimization time.  In particular, we tried to vary the hidden features from 128, 64, 32 to 16, and their corresponding model sizes are 137 KB, 53 KB, 23 KB, and 11 KB.  Figure~\ref{fig:model_size_PSNR}  shows a plot of the average PSNR computed over all these 200 test images for each model.  Results are shown for the entire image and out-of-gamut pixels only.   Figure~\ref{fig:visual-modelsize} shows qualitative results on  a test image for the different model sizes.  While the larger model gave a slightly better performance, the 23 KB model provided a comparable result, with memory size that can easily be included as a comment field in an sRGB image.   The model below 23 KB performed poorly. See supplemental materials for additional ablations.

\section{Discussion and concluding remarks}

We have presented a framework to recover wide-gamut color values lost due to the gamut reduction step applied when converting a ProPhoto image to a sRGB image. We cast our task as a restoration problem with the goal of restoring the original color loss due to gamut clipping. By integrating our approach into the gamut reduction step when the image is converted to sRGB, we have the opportunity to optimize a model directly against the known restoration target---a luxury most restoration problems such as deblurring and image sampling do not have. This allowed us to use a lightweight MLP network to predict the original color signal. Compared to the overall sRGB image size (typically 2--5 MB), the 23 KB memory overhead for the GamutMLP model is negligible and means that the color fidelity recovery is obtained virtually for free.

Our experiments show that per-image MLP optimization provides much better results than pre-trained DNN models for color recovery. Furthermore, our small GamutMLP provides comparable performance in PSNR compared with larger MLP-based neural implicit functions but requires significantly less memory size and optimization time (within 2 seconds).   As part of this effort, we have also generated a new dataset of 2200 images with high-color fidelity that will be useful in advancing research in this area. Our code and dataset can be found on the project website: \url{https://gamut-mlp.github.io}.

\paragraph{Acknowledgment} This study was funded in part by the Canada First Research Excellence Fund for the Vision: Science to Applications (VISTA) program, an NSERC Discovery Grant, and an Adobe Gift Award.

{\small
\bibliographystyle{ieee_fullname}
\bibliography{egbib}
}

\setcounter{section}{0}
\renewcommand\thesection{\Alph{section}}
\section{Mapping between color spaces}\label{sec:mapping}
For our Eq. (1) in the main paper, we convert ProPhoto images to sRGB images using a single $3\times 3$ matrix $\M$.  This conversion using $\M$ is comprised of three parts: the transformation from ProPhoto colors to the profile connection space (i.e., CIE XYZ) $\M^{-1}_{ProPhoto}$, the chromatic adaptation $\C_{CAT02}$, and the conversion from CIE XYZ to  sRGB with $\M_{sRGB}$.  The full matrix can be written as:
\begin{equation}
    \M = \M_{sRGB} \C_{CAT02} \M^{-1}_{ProPhoto}
\end{equation}

Different color spaces use different white points.  For example, sRGB and ProPhoto RGB use D65 and D50 respectively. If we want to transform ProPhoto colors to the sRGB space, first we convert ProPhoto colors to the profile connection space, then we chromatically adapt those tristimulus values to the white point of sRGB (D65), and finally convert them to sRGB. We choose the standard chromatic adaptation CIECAM02~\cite{ciecam02}. The complete transformation is:
\begin{equation}
    \M = \begin{bmatrix}
         2.0365 & -0.7376 & -0.2993 \\
        -0.2257 & 1.2232 & 0.0027 \\
        -0.0105 & -0.1349 & 1.1452 \\
         \end{bmatrix}
\end{equation}

\section{Training details for baseline methods and inference time }

The four baseline DNNs used different training sizes in their github repositories.  We felt to be fair we should use a common dataset when training all methods. The 512$\times$512 crop was the most common input size. The per-image optimization approaches were all significantly better than pre-trained DNNs, so small improvements based on different training sizes might not matter.

Inference time over the test dataset of our lightweight MLPs is around \textbf{0.27 seconds}, while it takes DNN-based methods
approximate \textbf{0.57 seconds} in average. The  ProPhoto-Sampled~\cite{Hoang_CIC2020} that uses a spatially varying mapping functions take up to 45 seconds per image.

We note that the MLP-based
methods purposely over-fit the MLP to the input image. Longer optimization gives better PSNR. However, to be practical we wanted the optimization to be within 2 seconds.

\section{Ablations}

\noindent{\textbf{Encoding function}}
The feature encoding (Eq. (4) in our main paper) used has been shown to be effective for implicit neural methods. It makes a big difference in our task. See Table~\ref{tab:mlp_encoding} below, where we show that our MLP with the encoding function is significantly better than the MLP without using the function. 

\begin{table}[h]
\centering
\caption{Our MLPs with and without using the encoding function.}
\resizebox{\columnwidth}{!}{%
\begin{tabular}{|l|c|c|c|c|}
\hline
\textbf{Method}                 & \multicolumn{1}{l|}{\textbf{RMSE$\downarrow$}} & \multicolumn{1}{l|}{\textbf{RMSE OG$\downarrow$}} & \multicolumn{1}{l|}{\textbf{PSNR$\uparrow$}} & \multicolumn{1}{l|}{\textbf{PSNR OG$\uparrow$}} \\ \hline
MLP (23KB) w/o enc. & 0.1467                             & 0.1547                                & 16.67                              & 16.21                                 \\ \hline
\rowcolor[HTML]{FFFC9E} 
MLP (23KB) w/ enc. & 0.0021                             & 0.0031                                & 53.65                              & 50.04                                 \\ \hline
\end{tabular}
}
\label{tab:mlp_encoding}
\end{table}

\section{Additional qualitative results}

\begin{figure*}[ht!]
  \vspace{-0.7cm}
  \includegraphics[width=0.86\textwidth]{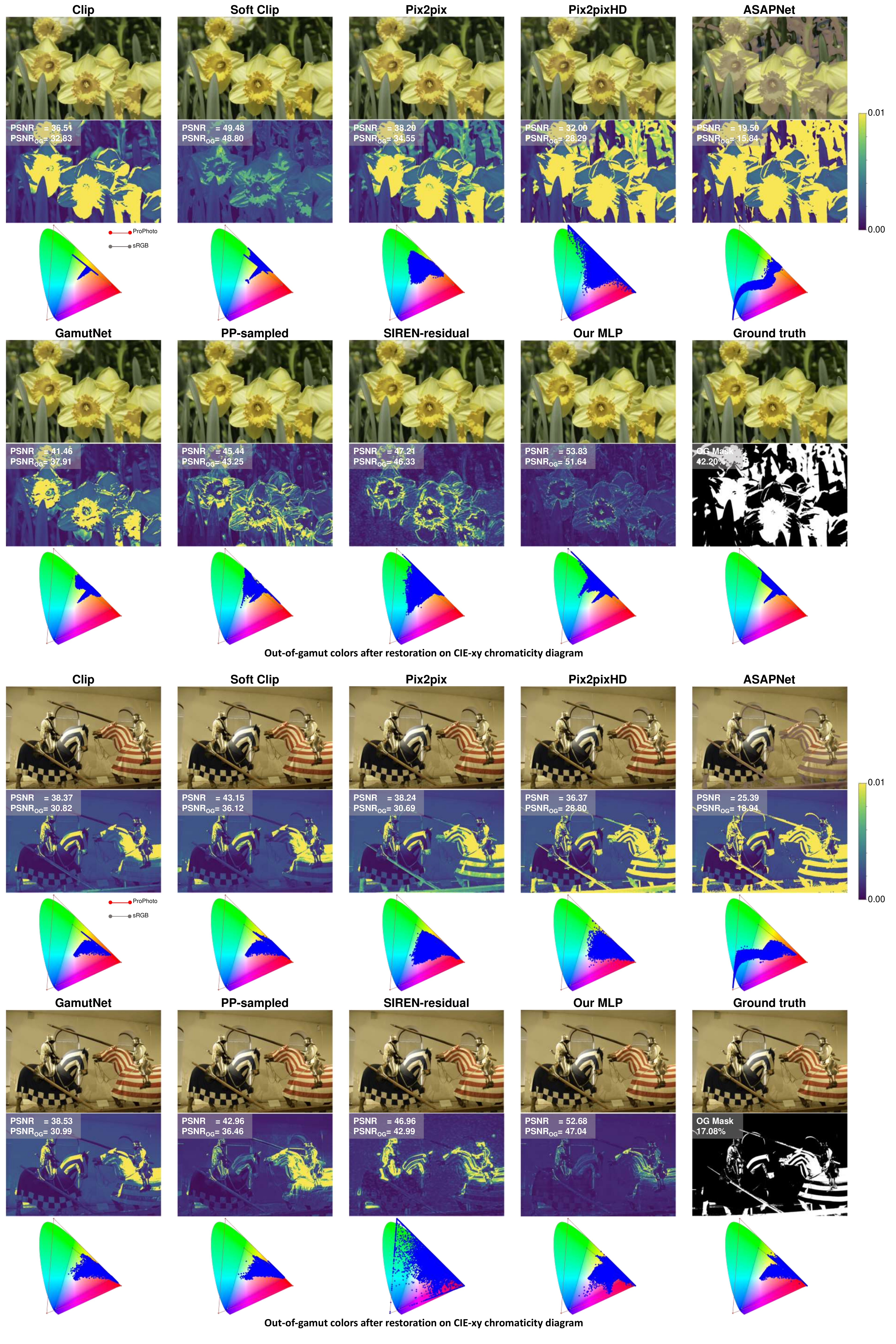}
  \centering
  \vspace{-0.4cm}
  \caption{Qualitative comparisons between the predicted ProPhoto full-size output of Clip, Soft Clip, Pix2pix~\cite{pix2pix}, Pix2PixHD~\cite{pix2pixHD}, ASAPNet~\cite{ASAPNet}, GamutNet~\cite{Hoang_CIC2021}, PP-sampled~\cite{Hoang_CIC2020}, SIREN~\cite{Sitzmann:2020:SIREN}-residual, and our optimized GamutMLP. Error maps of per-pixel RMSE and plots of out-of-gamut (OG) colors on CIE-xy chromaticity diagram with the gamuts of sRGB and ProPhoto are shown.}
  \label{fig:visual-results-supp-1}
  \vspace{-0.5cm}
\end{figure*}

\begin{figure*}[ht!]
  \vspace{-0.7cm}
  \includegraphics[width=0.86\textwidth]{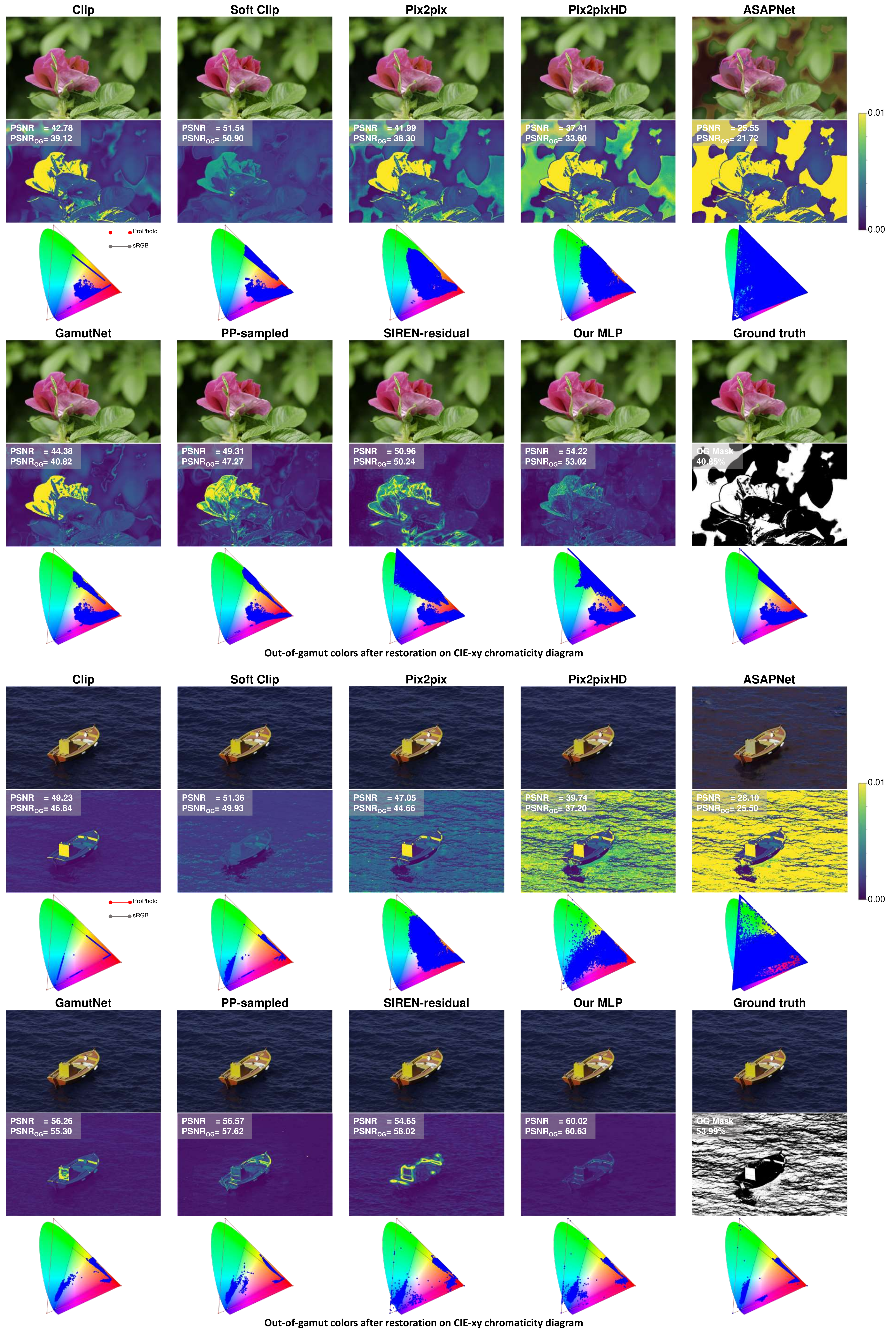}
  \centering
  \vspace{-0.4cm}
  \caption{Qualitative comparisons between the predicted ProPhoto full-size output of Clip, Soft Clip, Pix2pix~\cite{pix2pix}, Pix2PixHD~\cite{pix2pixHD}, ASAPNet~\cite{ASAPNet}, GamutNet~\cite{Hoang_CIC2021}, PP-sampled~\cite{Hoang_CIC2020}, SIREN~\cite{Sitzmann:2020:SIREN}-residual, and our optimized GamutMLP. Error maps of per-pixel RMSE and plots of out-of-gamut (OG) colors on CIE-xy chromaticity diagram with the gamuts of sRGB and ProPhoto are shown.}
  \label{fig:visual-results-supp-2}
  \vspace{-0.5cm}
\end{figure*}

We provide additional qualitative results as shown in the main paper. Figure~\ref{fig:visual-results-supp-1} and Figure~\ref{fig:visual-results-supp-2} compare our method with others on the test set. As shown in the figures, especially in per-pixel RMSE error maps, our approach achieve better qualitative results compared with others.

\end{document}